\documentclass[11pt]{article}

\usepackage[final]{acl}

\usepackage{times}
\usepackage{latexsym}

\usepackage[T1]{fontenc}
\usepackage[utf8]{inputenc}

\usepackage{microtype}

\usepackage{inconsolata}

\usepackage{graphicx}

\usepackage{amsmath}
\usepackage{amssymb}
\usepackage{multirow}
\usepackage{makecell}

\usepackage{booktabs}
\usepackage[table]{xcolor}
\usepackage{float}

\AtBeginDocument{
  \setlength{\abovedisplayskip}{5pt plus 1pt minus 1pt}     
  \setlength{\belowdisplayskip}{5pt plus 1pt minus 1pt}
  \setlength{\abovedisplayshortskip}{3pt plus 1pt minus 1pt}
  \setlength{\belowdisplayshortskip}{3pt plus 1pt minus 1pt}
}

\definecolor{mygreen}{rgb}{0.3, 0.7, 0.2}

\title{Intent Mismatch Causes LLMs to Get Lost in Multi-Turn Conversation}

\author{
 \textbf{Geng Liu\textsuperscript{1,2}},
 \textbf{Fei Zhu\textsuperscript{2}},
 \textbf{Rong Feng\textsuperscript{1,2}},
 \textbf{Changyi Ma\textsuperscript{3}},
 \textbf{Shiqi Wang\textsuperscript{1}},
 \textbf{Gaofeng Meng\textsuperscript{2}}\\
 {\normalsize \textsuperscript{1}College of Computing, City University of Hong Kong, Hong Kong SAR, China.} \\
 {\normalsize \textsuperscript{2}Centre for Artificial Intelligence and Robotics, HKISI, CAS} \\
 {\normalsize \textsuperscript{3}School of Artificial Intelligence, Jilin University, China} \\
 {\texttt{\small gengliu6@my.cityu.edu.hk, zhfei2018@gmail.com}}
}

\begin{document}
\maketitle
\begin{abstract}
Multi-turn conversation has emerged as a predominant interaction paradigm for Large Language Models (LLMs). Users often employ follow-up questions to refine their intent, expecting LLMs to adapt dynamically. However, recent research \cite{laban2025llms} reveals that LLMs suffer a substantial performance drop in multi-turn settings compared to single-turn interactions with fully specified instructions, a phenomenon termed ``Lost in Conversation'' (LiC). While this prior work attributes LiC to model unreliability, we argue that the root cause lies in an \textit{intent alignment gap} rather than intrinsic capability deficits.
In this paper, we first demonstrate that LiC is not a failure of model capability but rather a breakdown in interaction between users and LLMs. We theoretically show that scaling model size or improving training alone cannot resolve this gap, as it arises from structural ambiguity in conversational context rather than representational limitations. 
To address this, we propose to decouple intent understanding from task execution through a Mediator-Assistant architecture.
By utilizing an experience-driven Mediator to explicate user inputs into explicit, well-structured instructions based on historical interaction patterns, our approach effectively bridges the gap between vague user intent and model interpretation. Experimental results demonstrate that this method significantly mitigates performance degradation in multi-turn conversations across diverse LLMs.
\end{abstract}

\begin{figure*}[t]
\centering
  \includegraphics[width=\textwidth]{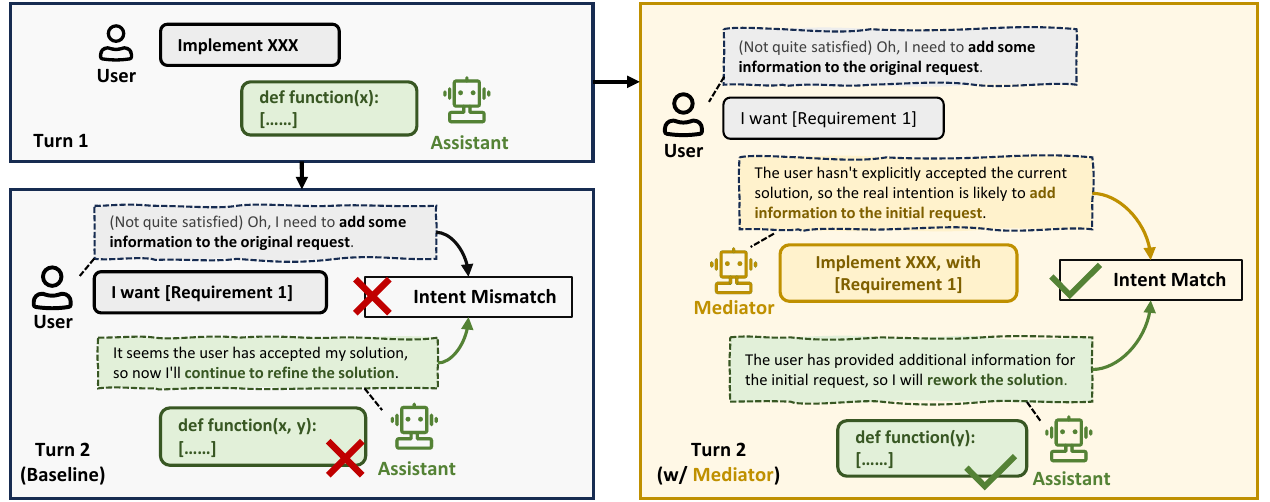}
  \caption{\textbf{Intent Mismatch in Multi-turn Dialogue.} (Left) The LiC benchmark simulates passive users who act as ``lazy'' interlocutors, omitting corrections for erroneous model assumptions. This behavior causes the Assistant's interpretation to progressively drift away from the user's true intent, leading to significant performance degradation. 
  (Right) Our approach introduces a Mediator to bridge this pragmatic gap by fundamentally decoupling intent inference from task execution. The Mediator aligns the Assistant with the user's true goals, effectively mitigating performance degradation.}
  \vspace{-10pt}
  \label{fig1}
\end{figure*}

\section{Introduction}\label{sec:intro}
In contemporary AI-assisted applications, multi-turn dialogue has become the primary mode of interaction between users and large language models (LLMs). Thanks to their massive parameter scales and extensive pretraining on diverse corpora, modern LLMs now exhibit impressive capabilities in language understanding, reasoning, and task execution, and they often perform remarkably well when given clear, complete, and well-structured instructions in a single turn. However, real-world user behavior rarely conforms to this idealized setting. In practice, users frequently start with vague, underspecified, or even internally inconsistent goals, and only gradually clarify and refine their true needs through an iterative conversational process with the model \cite{zamfirescu2023johnny, min2020ambigqa}. 
This incremental, exploratory nature of human problem formulation poses substantially greater challenges for LLMs than standard single-turn benchmarks: the model must not only understand and solve the current subtask, but also continually infer, update, and realign with a moving target of user intent across turns. 

Recent research \cite{laban2025llms} presents a set of controlled experiments designed to simulate the instruction underspecification that frequently occurs in human conversation \cite{herlihy2024overcoming,zipf2016human}. The study systematically compares performance under ``single-turn, fully specified'' (Full) versus ``multi-turn, underspecified'' (Sharded) interactions, revealing a substantial performance degradation of approximately 30\% for all evaluated LLMs. The authors argue that under incomplete information, LLMs tend to make premature assumptions early in the dialogue and subsequently ``lock in'' these assumptions, causing the final responses to drift away from the user's true intent. They term this phenomenon ``Lost in Conversation'' (LiC) and primarily attribute it to the reduced reliability of LLMs in multi-turn dialogue. On this basis, they advocate that LLMs should natively support multi-turn interaction and that model builders should jointly optimize models' aptitude and reliability in iterative conversational settings.

In this work, we revisit this phenomenon and offer a different explanatory perspective. We argue that:
(1) \textbf{Making early assumptions and providing tentative answers is not simply erroneous behavior, but a rational strategy induced by the dominant training objective of being helpful} \cite{ouyang2022training} and the penalty often associated with evasive responses in RLHF pipelines. Under conditions of incomplete information, the model is inclined to construct a plausible task formulation for a typical user and produce a provisional answer based on that formulation, instead of repeatedly refusing to answer or endlessly requesting additional information.
(2) \textbf{The primary bottleneck in failed multi-turn conversations is not a lack of model capacity or reasoning depth, but a pragmatic mismatch between user expression and model interpretation} (Figure~\ref{fig1} left). Users exhibit systematic individual variation, where the same utterance may map to disparate underlying intentions. General-purpose LLMs, aligned to the ``average'' user, fail to adapt to these idiosyncratic behaviors. For instance, models frequently misinterpret a user's fragmentary continuation as a confirmation of previous assumptions rather than a correction, thereby reinforcing an incorrect context.

To address this, we propose a framework that fundamentally decouples intent understanding from task execution. We operationalize this through a Mediator-Assistant pipeline, where a Mediator explicates user inputs to explicitly articulate latent requirements before they reach the execution Assistant. 
To align with specific user pragmatics, we employ an LLM-based Refiner to automatically distill explicit guidelines by analyzing the discrepancies between failed and successful interaction trajectories.
These guidelines then serve as context for the Mediator, enabling the system to bridge the alignment gap and adapt to individual user behaviors without the need for weight updates.

Our approach directly addresses the root cause of LiC: the misalignment between how users express intent and how models interpret it (Figure~\ref{fig1} right). By bridging this gap through adaptive input rewriting, we demonstrate substantial recovery of multi-turn performance across diverse LLMs, highlighting the critical role of user-aware intent modeling in conversational AI.

\section{Related Works}
\paragraph{Multi-turn Dialogue Evaluation.} Recent benchmarks for multi-turn dialogue, such as MT-Bench~\cite{zheng2023MTBench}, MT-Bench-101~\cite{bai2024mt}, and LOCOMO~\cite{maharana2024evaluating}, primarily focus on either (i) sequential task decomposition (e.g., planning a trip over multiple steps) or (ii) long-context retention in extended conversations. However, these settings often assume that each turn is sufficiently specified or that the full task context is available early in the dialogue. In contrast, our work targets a more challenging regime: \textit{incremental intent revelation}, where the user’s goal is only partially observable at each turn and may contradict earlier model assumptions—a scenario systematically studied in the “Lost in Conversation” (LiC) framework~\cite{laban2025llms} but largely overlooked by existing benchmarks.

\paragraph{Clarification and Intent Disambiguation.}
Another line of research encourages LLMs to actively seek clarification when faced with ambiguous queries~\cite{li2025verifiable, herlihy2024overcoming}. While effective in controlled settings, such approaches often conflict with real-world helpfulness norms, as users typically expect immediate, provisional responses rather than repeated clarification requests. As argued in Section~\ref{sec:intro}, premature assumption-making is a rational outcome of prevailing training objectives. Instead of modifying the model’s behavior, we preserve its default helpfulness while correcting misinterpretations upstream, through an adaptive mediator that refines inputs into unambiguous and complete instructions.

\paragraph{Personalization in LLMs.}
A growing body of work explores personalizing LLMs via parameter-efficient fine-tuning (PEFT)~\cite{xu2023}, user-specific adapters~\cite{zhong-etal-2021-useradapter}, or memory-augmented architectures. Systems like Mem0~\cite{mem0}, A-Mem~\cite{xu2025mem}, and MemoryBank~\cite{zhong2024memorybank} store user facts to enable long-term contextual awareness.
However, these approaches primarily address \textit{factual} personalization (e.g., remembering user preferences) rather than \textit{pragmatic} alignment, which is the challenge of interpreting ambiguous utterances according to a user’s idiosyncratic expression style.

\begin{figure*}[t]
  \includegraphics[width=\textwidth]{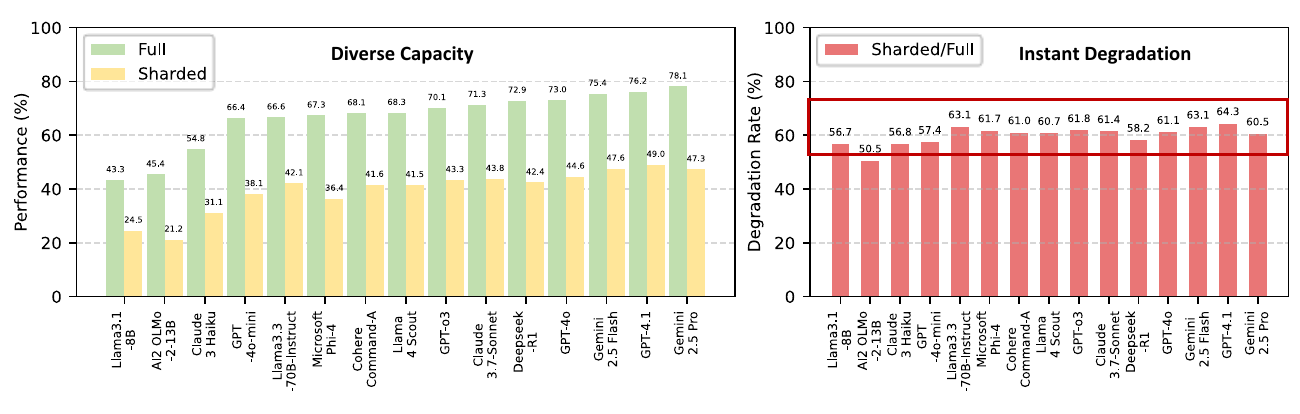}
  \vspace{-25pt}
  \caption{\textbf{Performance comparison across different LLMs on the LiC benchmark~\cite{laban2025llms}.} While absolute performance improves with model scale, the relative performance degradation remains strikingly constant ($\sim$60\%). This structural invariance suggests that the bottleneck lies in the alignment prior rather than model capacity.}
  \vspace{-10pt}
  \label{fig:scaling_invariance}
\end{figure*}

\section{Problem Analysis}
\label{sec:problem_analysis}
We formulate the multi-turn interaction between a user and an LLM assistant within a latent variable framework. Let $I_t \in \mathcal{I}$ denote the user's deep intent (the specific goal) at turn $t$, and $T \in \mathcal{T}$ represent the user's expression habits and pragmatic patterns.
In the $t$-th turn of interaction, the input to the LLM is the accumulated dialogue context $C_t$, consisting of the sequence of historical user utterances and assistant responses:
\begin{equation}
    C_t = (u_1, a_1, u_2, a_2, \dots, u_t).
\end{equation}
The user's current utterance $u_t$ is generated via a stochastic process $u_t \sim P_{\text{user}}(u \mid I_t, T, C_{t-1})$. Crucially, this process acts as a lossy projection, where complex, high-dimensional intents are compressed into low-dimensional and often ambiguous surface forms.
The LLM, defined by parameters $\theta$, aims to generate a response $R$ conditioned on the observed context:
\begin{equation}
    R \sim P_{\theta}(R \mid C_t).
\end{equation}

\subsection{Performance Decomposition}
We posit that the model's performance in multi-turn scenarios is not a monolithic metric but can be theoretically decomposed into two orthogonal components: (1) Intent Inference, the ability to recover the true intent $I_t$ from $C_t$; and (2) Task Execution, the ability to solve the identified intent $I_t$. 
Assuming that the true intent $I_t$ is a sufficient statistic for the task such that $R$ becomes conditionally independent of the noisy context $C_t$ given $I_t$ (i.e., $P_{\theta}(R \mid I_t)\approx P_{\theta}(R \mid I_t,C_t)$), we derive:
\begin{equation}
\begin{aligned}
    P_{\theta}(R \mid C_t) = \sum_{I_t \in \mathcal{I}} \underbrace{P_{\theta}(R \mid I_t)}_{\text{Execution}} \cdot \underbrace{P_{\theta}(I_t \mid C_t)}_{\text{Inference}}.
    \label{eq:decomposition}
\end{aligned}
\end{equation}

This decomposition reveals the fundamental mechanism behind performance degradation in multi-turn dialogues. 
The execution capability $P_{\theta}(R \mid I_t)$ represents the model's intrinsic reasoning ability given a perfectly defined instruction. This is largely determined by the model's pre-training and remains relatively stable for a specific task.
However, the accuracy of intent inference $P_{\theta}(I_t \mid C_t)$ faces severe challenges as interactions progress. In an ideal single-turn setting, users tend to provide self-contained descriptions, making the intent $I_t$ clearly inferable from the utterance. Conversely, in multi-turn dialogues, driven by the \textit{principle of least effort} and individual habits ($T$), users often generate highly personalized, ambiguous, and fragmented surface forms (e.g., using pronouns or vague directives) based on the same intent $I_t$. This pragmatic ellipsis significantly widens the semantic gap between the surface utterance and the deep intent. 
Consequently, the LLM fails not because it loses the capability to solve the problem, but because it cannot penetrate the user's ambiguous expression to clearly define the problem. 

\subsection{The Information Bottleneck}
The challenge of maximizing $P(I_t \mid C_t)$ is not merely a lack of model capacity, but an information-theoretic limit. 
Mathematically, the conditional entropy of the user's intent given the context, $H(I_t \mid C_t)$, remains high because the mapping from intent to utterance is many-to-one. When critical constraints are omitted by the user due to lossy compression, the missing information bits are simply absent from $C_t$.

Under such high uncertainty, a frozen LLM $P_\theta$ tends to revert to its training priors. It implicitly solves for $\arg\max_{I_t} P_{\text{pretrain}}(I_t \mid C_t)$, aligning with the ``average user'' rather than the specific individual. This manifests as generic convergence: the model assumes the most statistically probable intent, leading to the LiC phenomenon where the model confidently answers the wrong question. 
Scaling the model size ($\theta$) does not solve this; it merely allows the model to better fit the prior distribution of the average user, essentially reinforcing the misalignment. When contextual uncertainty is high, a frozen LLM defaults to its pretraining prior, generating $I_t$ that maximizes (approximately) $P_\theta(I_t \mid C_t)$. Since this distribution encodes population-level patterns rather than individual preferences, the output often reflects a stereotypical “average user,” leading to confidently incorrect responses (LiC). Scaling model size only sharpens this prior, exacerbating the misalignment in the absence of personalized signals.

This theoretical limitation is corroborated by empirical evidence. As illustrated in Figure~\ref{fig:scaling_invariance}, a re-analysis of experimental results from \citet{laban2025llms} reveals a striking pattern: while stronger models achieve higher absolute scores, the \textit{relative performance degradation} between fully specified and underspecified settings remains remarkably constant (approximately 60\%) across diverse model sizes and families.
We attribute this \textbf{invariant degradation} to the homogeneity of the alignment prior. Although models differ in capacity $\theta$, they are predominantly pre-trained and aligned on similar vast corpora, leading them to converge towards a shared representation of the ``average user'' ($P_{\text{avg}}$). In the absence of specific constraints, all models fall back to this shared prior:
\begin{equation}
    \arg\max_{I_t} P_{\theta}(I_t \mid C_{t}) \approx \arg\max_{I_t} P_{\text{avg}}(I_t \mid C_{t}).
\end{equation}
Since the specific user intents in the benchmark deviate from this population mean in a fixed manner, the divergence between the actual user intent and the average prior acts as a constant structural penalty. Consequently, merely scaling model parameters optimizes the fit to $P_{\text{avg}}$ but does not bridge the semantic gap to the specific individual, rendering the ambiguity strictly unresolvable via scaling alone.

\begin{figure*}[t]
  \includegraphics[width=\textwidth]{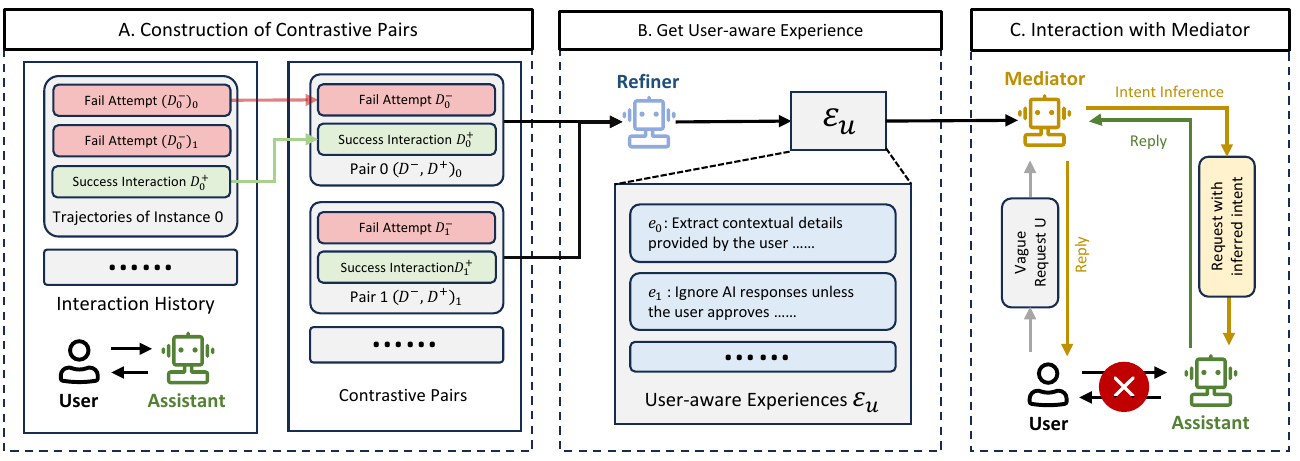}
  \caption{\textbf{Pipeline of the Mediator Framework.} We construct contrastive pairs by extracting a failed conversational trajectory $D^{-}$ and the corresponding successful trajectory $D^{+}$ for the same task instance from the user's historical logs. The Refiner distills these pairs into explicit pragmatic experiences $\mathcal{E}$, which guide the Mediator to explicate ambiguous user contexts into precise instructions for the Assistant. The Mediator operates as a transparent alignment layer, decoupling the user from the raw execution model while maintaining a seamless interaction flow.
}
\vspace{-10pt}
  \label{fig2}
\end{figure*}

\subsection{Reducing Entropy via History}
The decomposition in Eq.~\eqref{eq:decomposition} implies that to improve performance, we must sharpen the distribution $P(I_t \mid \dots)$. Since the entropy $H(I_t \mid C_t)$ is constrained by the information loss in $C_t$, we must introduce an auxiliary variable: the user's generalized interaction history, denoted as $\mathcal{H}$.

We propose that while $C_t$ is ambiguous on its own, the extended context $(C_t, \mathcal{H})$ contains sufficient information to recover $I_t$. 
By encapsulating longitudinal evidence of user behavior, including past dialogue trajectories and interaction outcomes, $\mathcal{H}$ implicitly encodes the user's specific expression habits $T$ and effectively acts as the ``compression key''.

Our method introduces a Mediator $M$ to approximate this inference process. Instead of asking the general model to guess $I_t$ solely from $C_t$, the Mediator computes:
\begin{equation}
    \hat{U} \sim P(U \mid C_t, \mathcal{H}).
\end{equation}
Here, $\hat{U}$ is a reconstructed, fully-specified instruction that acts as a proxy for the latent intent $I_t$. By conditioning on $\mathcal{H}$, the Mediator significantly reduces the conditional entropy: 
\begin{equation}
    H(I_t \mid C_t, \mathcal{H}) \ll H(I_t \mid C_t).
\end{equation}
The downstream LLM then operates on this low-entropy input: $R \sim P_{\theta}(R \mid \hat{U})$. This effectively bypasses the information bottleneck, allowing the general model to execute tasks with the clarity and precision of a single-turn interaction.

\section{Method}
\label{sec:method}

Building upon the analysis in \S\ref{sec:problem_analysis}, we propose a Mediator-Assistant framework designed to resolve the pragmatic mismatch in multi-turn dialogues. 
To bridge the information gap, we leverage the user's generalized history $\mathcal{H}$. In our implementation, $\mathcal{H}$ consists of raw contrastive interaction pairs (successful vs. failed trajectories). However, raw history could be noisy and high-dimensional. Therefore, we introduce a Refiner module (R) to distill $\mathcal{H}$ into a compact set of explicit Experiences ($\mathcal{E}$). These experiences serve as the ``pragmatic profile'' for the Mediator $M$.
Our approach adopts a training-free, experience-driven paradigm. This strategic choice ensures immediate adaptability: the system learns from history without parameter updates, bypassing the storage and versioning overheads of per-user fine-tuning.

\subsection{The Mediator Framework}
\label{subsec:framework}
The core premise, as derived in Eq.~\eqref{eq:decomposition}, is that the raw context $C_t$ acts as a lossy compression of the true intent $I_t$. Direct interaction fails because the generic Assistant ($A$) maximizes $P(R\mid C_t)$ based on population priors rather than individual intent. The Mediator acts as an alignment layer, approximating the inference distribution by conditioning on the distilled experiences: $P(I_t\mid C_t, \mathcal{E})$.

The Mediator takes the accumulated context $C_t$ and experiences $\mathcal{E}$.
It analyzes the discrepancy between the surface utterance and the latent intent, generating a reconstructed instruction $\hat{U}$. This $\hat{U}$ is an explicit, fully specified articulation of $I_t$. The Assistant $A$ then generates the response based on this low-entropy input:
\begin{equation}
R = A(\hat{U}),\quad \text{where } \hat{U} \approx I_t.
\end{equation}

For a specific user, the distilled experiences are aggregated into a fixed knowledge base $\mathcal{E}_u$, which will be injected into the Mediator as a system instruction.
Mathematically, the Mediator performs the mapping:
\begin{equation}
\hat{U} = M(C_t \mid \mathcal{E}_u).
\end{equation}
During inference, the Mediator references this established profile to interpret the current context $C_t$. Since $\mathcal{E}_u$ explicitly encodes the user's specific pragmatic habits and constraints, $M$ can accurately detect omitted information in the surface utterance and synthesize the clarified prompt $\hat{U}$.

\subsection{Experience Acquisition}
\label{subsec:acquisition}
This subsection details how we construct the raw history $\mathcal{H}$ and how the Refiner transforms it into actionable experiences $\mathcal{E}$.

\paragraph{Contrastive Pair Construction.}
We premise our data construction on a pervasive behavioral pattern in human-computer interaction: iterative query reformulation \cite{huang2009analyzing, odijk2015struggling}. Real-world users typically exhibit goal-directed persistence; when an initial ambiguous utterance fails to elicit the desired response, users rarely abandon the task immediately. Instead, they engage in a trial-and-error process, refining their instructions until the model successfully executes the task.
This behavior naturally yields a dataset of paired trajectories within successful sessions. We view the final, successful turn as the ground-truth articulation ($D^{+}$) of the user's intent, while the preceding failed interaction sequence serves as the ambiguous context ($D^{-}$).
We define the generalized history $\mathcal{H}$ as a collection of contrastive interaction pairs $(D^{-},D^{+})$. 
From an information-theoretic perspective, the difference between $D^{-}$ and $D^{+}$ explicitly reveals the latent information bits that were omitted in the ambiguous context but are required to reduce the conditional entropy $H(I_t\mid C_t)$.

To operationalize this framework within our simulation benchmark, we construct these pairs synthetically using interaction logs. 
We identify instances where the model exhibits performance discrepancies: specifically, instances where the model fails in the multi-turn conversational setting but succeeds when the same task is presented in a single-turn format.
From these instances, we construct discrete contrastive pairs. For each task selected for experience mining, we sample a single specific failed conversational trajectory to serve as $D^-$ and pair it directly with the corresponding successful single-turn input as $D^+$. We do not aggregate all possible failure modes for a task; instead, we establish a one-to-one mapping between a specific noisy history and an effective input. In this context, the successful single-turn prompt acts as a proxy for the user's ``final successful turn.'' By analyzing this specific pair, the system can learn to bridge the gap between particular context and effective intent.

\paragraph{Experience Distillation via Refiner.}
To extract explicit inference rules from these noisy interaction logs, we employ a Refiner ($R$). The Refiner performs an inductive analysis on the contrastive pairs. It receives contrastive pairs $(D^{-}, D^{+})$ and performs a contrastive analysis: what underlying pattern can be learned to identify user's true intents from the trajectories in the future?
The output is a set of structured textual guidelines $\mathcal{E}=\{e_0,e_1,...,e_n\}$. These guidelines do not merely memorize the specific task content, but distill the pragmatic strategy (e.g., ``If the user has not explicitly approved the previous solution, he is not satisfied with it.''). These distilled experiences serve as the context for the Mediator.

\begin{table*}[]
\centering
\setlength{\tabcolsep}{7pt}
\small
\begin{tabular}{llcccccccccc}
\toprule
\multirow{2}{*}{Model} & \multirow{2}{*}{Method} & \multicolumn{2}{c}{Code} & \multicolumn{2}{c}{Database} & \multicolumn{2}{c}{Actions} & \multicolumn{2}{c}{Math} & \multicolumn{2}{c}{Average} \\ \cmidrule{3-12} 
&          & $\bar{P}$ & $R$ & $\bar{P}$ & $R$ & $\bar{P}$ & $R$&$\bar{P}$ & $R$ &$\bar{P}$ & $R$  \\ 
\midrule
\cellcolor{white}\multirow{4}{*}{GPT-4o-mini} 
&  Full    & 74.2 & 76.0 & 92.5 & 93.5 & 93.7 & 92.4 & 87.2 & 70.9 & 86.9 & 83.2 \\
\cmidrule{2-12}
&  Sharded & 51.4 & 57.0 & 52.5 & 54.2 & 45.5 & 60.0 & 64.9 & 45.6 & 53.6 & 54.2 \\
&  w/ Ours & 66.9 & 63.6 & 65.3 & 59.8 & 85.7 & 81.2 & 77.7 & 70.4 & 73.9 & 68.8 \\ 
&  Gain & \textcolor{mygreen}{+15.5} & \textcolor{mygreen}{+6.6} & \textcolor{mygreen}{+12.8} & \textcolor{mygreen}{+5.6} & \textcolor{mygreen}{+40.2} & \textcolor{mygreen}{+21.2} & \textcolor{mygreen}{+12.8} & \textcolor{mygreen}{+24.8} & \textcolor{mygreen}{+20.3} & \textcolor{mygreen}{+14.6} \\ 
\midrule
\cellcolor{white}\multirow{4}{*}{GPT-5.2}  
&  Full    & 83.2 & 84.0 & 96.3 & 95.9 & 90.2 & 93.2 & 94.5 & 89.4 & 92.7 & 85.4 \\
\cmidrule{2-12}
&  Sharded & 39.4 & 44.0 & 49.4 & 48.0 & 35.6 & 46.6 & 69.6 & 48.6 & 48.5 & 46.8 \\
&  w/ Ours & 69.1 & 70.1 & 64.5 & 56.7 & 76.2 & 65.2 & 80.6 & 62.0 & 72.6 & 63.5 \\ 
&  Gain & \textcolor{mygreen}{+29.7} & \textcolor{mygreen}{+26.1} & \textcolor{mygreen}{+15.1} & \textcolor{mygreen}{+8.7} & \textcolor{mygreen}{+40.6} & \textcolor{mygreen}{+18.6} & \textcolor{mygreen}{+11.0} & \textcolor{mygreen}{+13.4} & \textcolor{mygreen}{+24.1} & \textcolor{mygreen}{+16.7} \\ 
\midrule
\cellcolor{white}\multirow{4}{*}{\makecell[l]{DeepSeek-\\v3.2-Thinking}} 
&  Full    & 98.3 & 95.9 & 94.4 & 88.8 & 92.2 & 88.6 & 94.0 & 81.6 & 94.7 & 88.7 \\
\cmidrule{2-12}
&  Sharded & 78.4 & 65.7 & 43.6 & 54.2 & 42.3 & 48.6 & 78.8 & 56.3 & 60.8 & 56.2 \\
&  w/ Ours & 86.1 & 84.8 & 67.3 & 55.9 & 88.0 & 71.6 & 86.3 & 67.3 & 81.9 & 69.9 \\ 
&  Gain & \textcolor{mygreen}{+7.7} & \textcolor{mygreen}{+19.1} & \textcolor{mygreen}{+23.7} & \textcolor{mygreen}{+1.7} & \textcolor{mygreen}{+45.7} & \textcolor{mygreen}{+23.0} & \textcolor{mygreen}{+7.5} & \textcolor{mygreen}{+11.0} & \textcolor{mygreen}{+21.1} & \textcolor{mygreen}{+13.7} \\ 
\bottomrule
\end{tabular}
\caption{\textbf{Main Results.}  Comparison of average performance ($\bar{P}$) and reliability ($R$) across three LLM backbones. All experiments are averaged over 5 runs. $R$ is calculated as the mean of $1 - (S_{\text{max}} - S_{\text{min}})$ across all instances.
\textit{Full} represents ideal instructions, while \textit{Sharded} represents ambiguous user inputs. \textit{w/ Ours} denotes the performance when using our Experience-Driven Mediator. The best improvements are highlighted in green.}
\vspace{-10pt}
\label{table_1}
\end{table*}

\section{Experiments}
To systematically evaluate the robustness of LLMs in multi-turn interactions and validate our proposed framework, we conduct extensive experiments leveraging the simulation benchmark introduced by \citet{laban2025llms}. We benchmark a diverse set of state-of-the-art models, including the widely-used GPT-4o-mini~\cite{hurst2024gpt}, the highly capable GPT-5.2, and the reasoning-specialized DeepSeek-V3.2-Thinking~\cite{deepseekai2024deepseekv32}. Details of experiment setup are available at \S\ref{sec:detail}.
In this section, we present our main results, demonstrating the universality of the LiC phenomenon and the efficacy of our Mediator across different model architectures (\S\ref{subsec:main_results}). We also provide an in-depth ablation analysis to distinguish pragmatic alignment from simple factual memory retrieval (\S\ref{subsec:ablation}).

\subsection{Main Results}
\label{subsec:main_results}
Table~\ref{table_1} presents the comprehensive evaluation results across four domains: Code, Database, Actions, and Math.
We compare three settings: 
(1) \textbf{Full}: The idealized upper bound, where users provide complete, unambiguous instructions in a single turn.
(2) \textbf{Sharded}: The baseline setting, representing simulated multi-turn conversations where context is fragmented and ambiguous.
(3) \textbf{w/ Ours}: The proposed framework, where the Experience-Driven Mediator reconstructs the \textit{Sharded} input into a specified instruction before passing it to the Assistant.
To rigorously assess model robustness against the stochasticity inherent in multi-turn generation, we conducted five independent runs for every experimental instance. We report two key metrics:
(1) Average Performance ($\bar{P}$): The mean score across the five runs.
(2) Reliability ($R$): A consistency metric calculated at the instance level. For each specific instance, we measure the divergence between its best and worst outcomes across the five runs $(S_\text{max}-S_\text{min})$. The reported $R$ is the average of $1-(S_\text{max}-S_\text{min})$ across all instances. 

\paragraph{The Persistence of LiC.}
Comparing the \textit{Full} and \textit{Sharded} settings reveals a severe alignment gap that transcends model architectures. First, standard instruction-tuned models exhibit dramatic degradation; for instance, \textbf{GPT-5.2} drops from a near-perfect $92.7\%$ to $48.5\%$. This confirms that scaling model capabilities alone cannot resolve the ambiguity problem. In the absence of explicit intent, even the most capable models inevitably resort to ungrounded guessing to fill the information gap, resulting in performance collapses similar to those observed in smaller baselines.
Crucially, even reasoning-enhanced models like \textbf{DeepSeek-v3.2-Thinking} fail to overcome this hurdle. Contrary to the expectation that intrinsic Chain-of-Thought (CoT) might infer missing context, DeepSeek's performance in the \textit{Sharded} setting remains severely compromised.
This result exposes a critical limitation: reasoning capabilities are ineffective against information ambiguity. While CoT excels at logical deduction when premises are clear, it cannot ``reason'' its way out of an information vacuum. Without the explicit context reconstruction provided by our Mediator, the model is forced into ungrounded speculation, regardless of its reasoning depth.

\paragraph{Efficacy of Mediator.}
Implementing our proposed Mediator yields substantial and consistent improvements across all backbones and domains.
On average, our method recovers performance by approximately 20\% in $\bar{P}$ and 15\% points in $R$.
Notably, even for the reasoning-enhanced DeepSeek-v3.2-Thinking, our approach secures a 21.1\% increase, demonstrating that the Mediator provides value complementary to advanced internal reasoning.
By reconstructing ambiguous contexts into explicit, self-contained instructions ($\hat{U}$), it collapses the uncertainty space for each individual instance. This ensures that the Assistant receives a low-entropy input, leading to consistently high-quality execution regardless of the random seed, effectively converting a probabilistic guessing game into a deterministic reasoning task.

\paragraph{Correlation between Reliability and Performance.}
We observe substantial improvements in both Performance ($\bar{P}$) and Reliability ($R$). This parallel growth demonstrate that multi-turn generation is not inherently prone to high variance. Instead, the data suggests that the previously observed instability stemmed primarily from intent ambiguity rather than intrinsic model stochasticity. By effectively aligning the Assistant with the intended constraints, our Mediator proves that reliability can be enhanced just as effectively as performance. This confirms that once the intent is grounded, the model's generation process transitions from a volatile guessing game to a robust and reproducible execution.

\subsection{Intent Alignment vs. Factual Memory}
\label{subsec:ablation}
To identify the root cause of the performance degradation in multi-turn dialogues, we compare our approach against two representative context utilization strategies (Table~\ref{tab:ablation_memory}): a naive summarization method (\textbf{w/ Sum}) and a mainstream RAG-based memory framework \textbf{Mem0} to persist user facts.

\begin{table}[]
\centering
\setlength{\tabcolsep}{6pt}
\begin{tabular}{m{1.5cm}ccccc}
\toprule
Method & C & D & A & M & Avg. \\
\midrule
% \rowcolor{gray!20}
Full    & 74.2 & 92.5 & 93.7 & 87.2 & 86.9 \\
\midrule
Sharded & 51.4 & 52.5 & 45.5 & 64.9 & 53.6 \\
w/ Sum & 57.1 & 44.9 & 54.6 & 62.0 & 54.7 \\
% Gain & \textcolor{mygreen}{+5.7} & \textcolor{red}{-7.6} & \textcolor{mygreen}{+9.1} & \textcolor{red}{-2.9} & \textcolor{mygreen}{+1.1} \\
w/ Mem0 & 52.8 & 56.9 & 49.4 & 66.9 & 56.5 \\
%Gain & \textcolor{mygreen}{+1.4} & \textcolor{mygreen}{+4.4} & \textcolor{mygreen}{+3.9} & \textcolor{mygreen}{+2.0} & \textcolor{mygreen}{+2.9} \\
w/ Ours & 66.9 & 65.3 & 85.7 & 77.7 & 73.9 \\
%Gain & \textcolor{mygreen}{+15.5} & \textcolor{mygreen}{+12.8} & \textcolor{mygreen}{+40.2} & \textcolor{mygreen}{+12.8} & \textcolor{mygreen}{+20.3} \\
\bottomrule
\end{tabular}
\caption{\textbf{Comparative Analysis.} We compare our method against Summarization (Sum) and a RAG-based memory framework (Mem0) on GPT-4o-mini. The marginal gains of Mem0 indicate that simply recalling factual context is insufficient for multi-turn robustness. In contrast, our method succeeds by explicitly reconstructing the user's intent.}
\vspace{-15pt}
\label{tab:ablation_memory}
\end{table}

\paragraph{Memory is not Understanding.}
A prevailing hypothesis suggests that models fail because they forget constraints or details from previous turns. If this were true, Mem0 should significantly restore performance by retrieving relevant historical facts. However, the results in Table~\ref{tab:ablation_memory} contradict this assumption. \textbf{w/ Mem0} yields only marginal gains over the baseline (53.6\% $\to$  56.5\%), and \textbf{w/ Sum} shows high variance, even degrading performance in some domains.

This finding highlights a critical distinction: retrieving context is not equivalent to resolving intent.
In the \textit{Sharded} setting, the model usually has access to the raw information (facts); the failure stems from its inability to determine how the user intends to apply those facts to the current ambiguous request. For example, Mem0 might successfully recall that ``the user wants a Python script,'' but it fails to clarify which specific logic from the conversation history should be applied now.

In contrast, by explicitly synthesizing an unambiguous instruction, our Mediator bridges the gap between raw context and actionable intent. This results in a decisive improvement, outperforming Mem0 by 17.4\%. This confirms that simply ``remembering'' is insufficient to solve the LiC problem, the system should explicitly align with the user's intent.

\begin{figure}[t]
\centering
  \includegraphics[width=0.9\columnwidth]{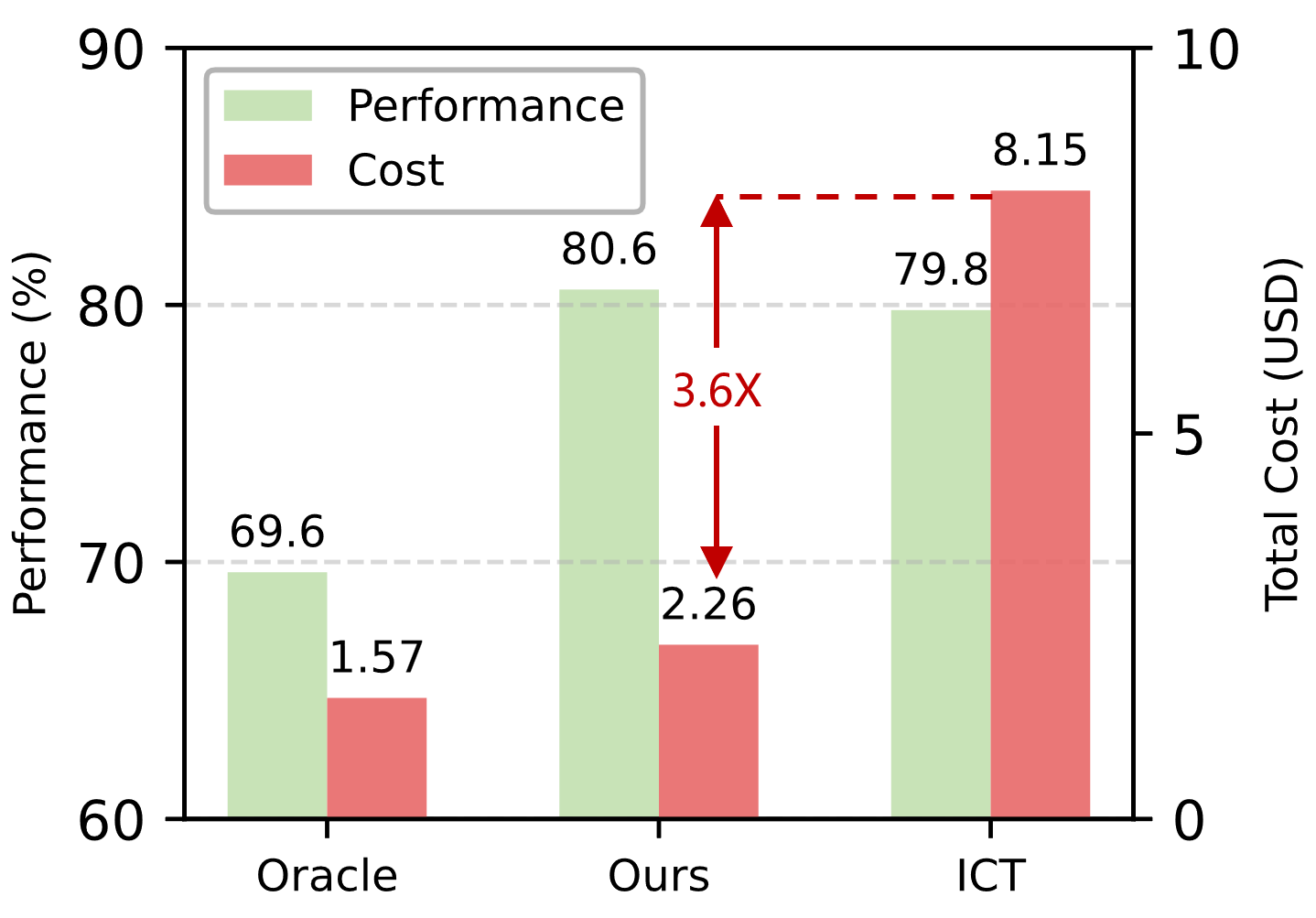}
  \caption{\textbf{Comparison with In-Context Learning.} We compare our method against the Oracle baseline and a Direct ICL approach. Our method delivers a substantial performance boost over the Oracle with only a marginal cost increase. While ICL achieves comparable accuracy to Ours, it consumes $3.6 \times$ more tokens.}
  \vspace{-10pt}
  \label{fig3}
\end{figure}

\paragraph{Why Refiner?}
We further investigate the necessity of the Refiner module by comparing it with an In-Context Learning (ICL) baseline, which embeds raw contrastive pairs directly into the Mediator's prompt, and the Oracle baseline. As shown in Figure~\ref{fig3}, our method significantly outperforms the Oracle while incurring only a negligible increase in token usage. In contrast, direct ICL achieves comparable accuracy to our method, but it incurs a $3.6 \times$ increase in token consumption due to verbose context. This demonstrates that Refiner distills verbose interaction logs into concise guidelines, ensuring practical inference efficiency without compromising performance.

\section{Conclusions}
In this paper, we revisited the ``Lost in Conversation'' (LiC) phenomenon, identifying it not as a fundamental deficit in model capability, but as an intent alignment gap between user expression and model understanding. We theoretically demonstrated that simply scaling up LLMs is insufficient to resolve this issue, necessitating an architectural intervention. To this end, we proposed a Mediator-Assistant framework equipped with an experience Refiner. By distilling historical contrastive trajectories into concise guidelines, our method enables the Mediator to explicate ambiguous inputs into explicit instructions, then significantly mitigates performance degradation in multi-turn interactions.

\section{Limitations}
Constrained by the limited scale of existing benchmarks, our current Refiner operates in a few-shot, non-parametric manner, extracting explicit and heuristic-level guidelines. While this design ensures efficiency, it captures only coarse-grained interaction patterns. Future work could leverage larger-scale datasets to transition towards parameterized training, enabling the Mediator to internalize more nuanced alignment strategies via fine-tuning rather than relying solely on in-context summaries.
Furthermore, the multi-turn settings in current benchmarks exhibit relatively homogeneous user logic. Developing more comprehensive benchmarks that mirror complex user behaviors remains a critical direction to further validate and evolve our framework.

\bibliography{custom}

\appendix

\section{Benchmark Details}
\label{sec:detail}

\subsection{Dataset Details}
We utilize the pre-constructed conversational data provided by \citet{laban2025llms}. The original benchmark encompassed six diverse tasks, which can be categorized into two distinct types based on their evaluation criteria. The first category, Binary Correctness Tasks, consists of Code, Database, Actions, and Math. For these tasks, success is defined by strict execution accuracy or exact constraint satisfaction. The second category, Refinement Tasks, comprises Data-to-text and Summarization, where the performance is evaluated based on generation quality rather than a single correct answer.

To transform these standard single-turn benchmarks into multi-turn conversational trajectories, the authors employed a semi-automated ``sharding'' pipeline. This process involved using a teacher LLM to decompose a fully specified instruction into a sequence of atomic, self-contained constraints (shards), which were subsequently rewritten to mimic a user progressively clarifying requirements. These shards were rigorously validated via human annotation to ensure semantic equivalence to the original instruction.

\subsection{Experimental Adaptations}
In this work, we introduce two critical adaptations to the original experimental setup to strictly isolate conversational reasoning failures from other confounding factors.

\paragraph{Task Selection.} 
First, we refine the scope of evaluation by focusing exclusively on the Binary Correctness Tasks (Code, Database, Actions, and Math). We explicitly exclude the Refinement Tasks (Data-to-text and Summarization) from our experiments. Our preliminary analysis indicated that LLMs often exhibit floor-level performance on these open-ended tasks even in single-turn baselines. Furthermore, the high variance inherent in the evaluation metrics for text generation complicates the analysis, making it difficult to distinguish between multi-turn context failures and general generation capability issues.

\paragraph{Sequential Sharding Strategy.} 
Second, we introduce a strict control regarding information ordering. The methodology in \citet{laban2025llms} randomly shuffled the order in which shards were revealed to the model. We argue that randomizing constraints often introduces artificial logical incoherence (e.g., modifying a variable before it is defined), which conflates the model's reasoning capability with its ability to handle ill-posed logical puzzles. To mitigate this, we enforce a sequential sharding strategy where we re-order the provided shards to follow a natural, logical progression (e.g., Problem Definition followed by specific constraints). This ensures that any observed performance degradation is strictly attributable to the model's inability to manage multi-turn context, rather than confusion caused by disordered inputs.

\paragraph{Few Shot Learning Set.} 
The current dataset has a limited amount of data. To minimize the data required for few-shot experience learning, we randomly select 5 instances from each task to construct historical dialogues. These instances are then removed from the test set to ensure a fair comparison.

\clearpage
\section{Prompts}
\begin{figure}[H]
\centering
  \includegraphics[width=\textwidth]{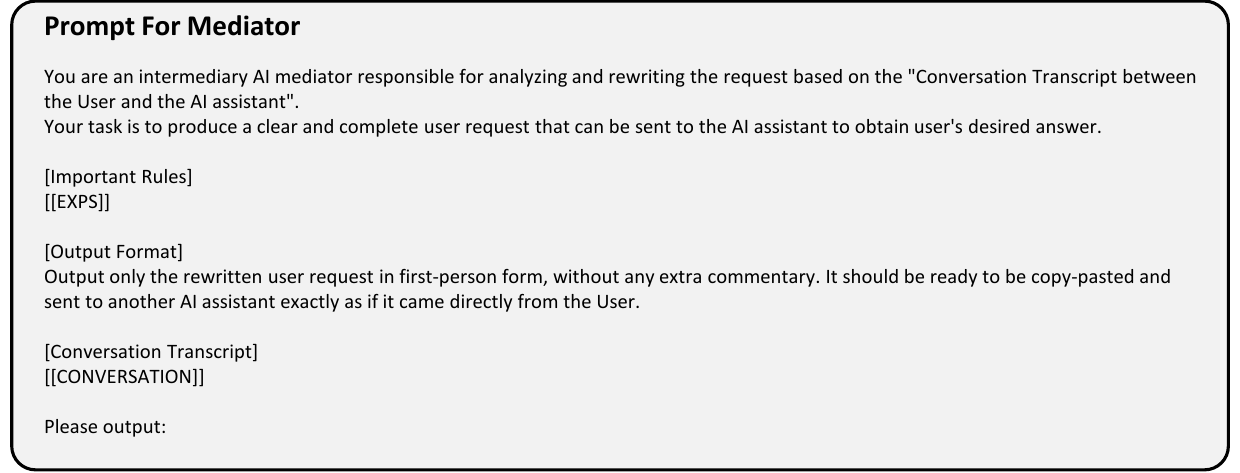}
  \label{p1}
\end{figure}

\begin{figure}[H]
\centering
  \includegraphics[width=\textwidth]{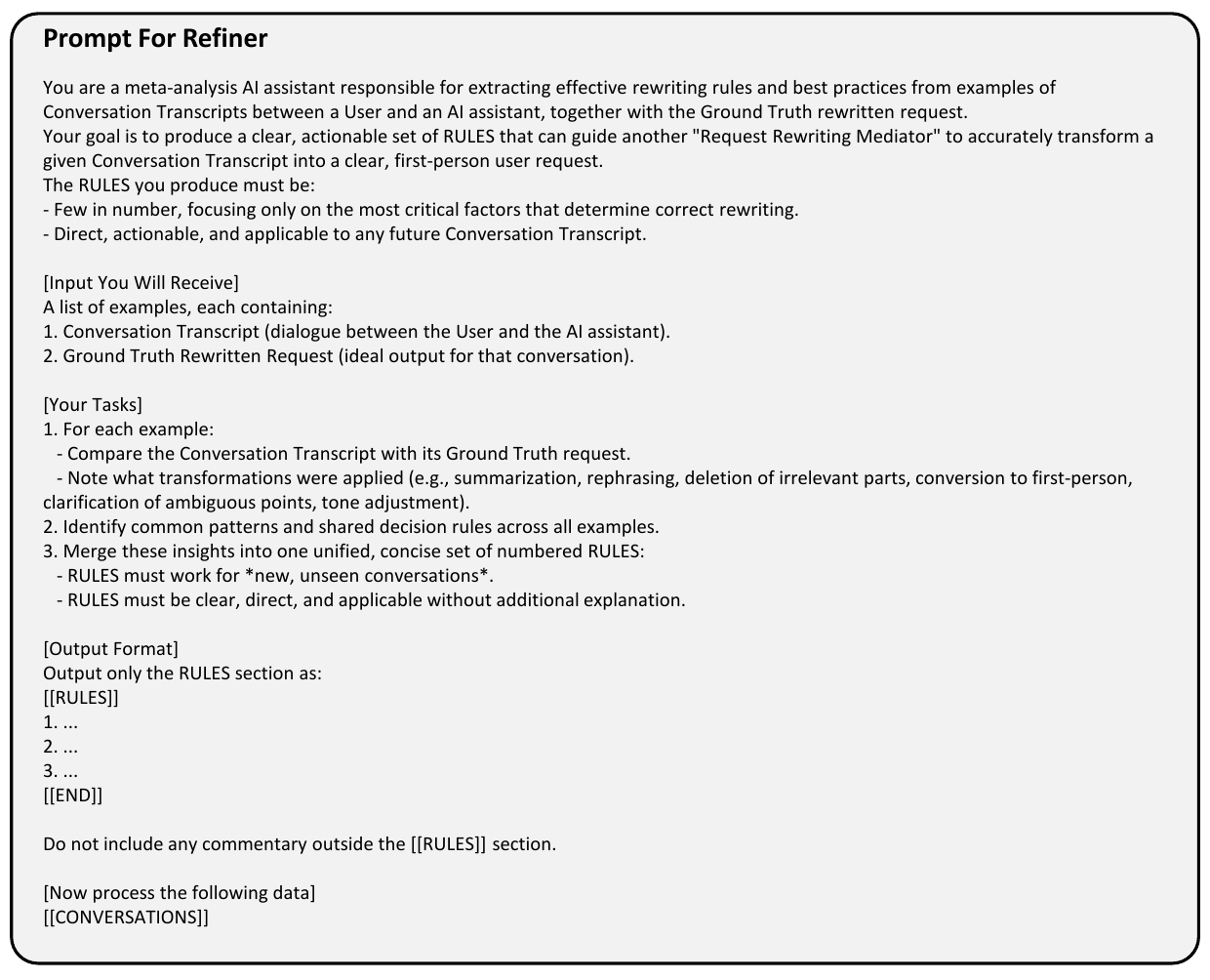}
  \label{p2}
\end{figure}

\clearpage
\begin{figure}[H]
\centering
  \includegraphics[width=\textwidth]{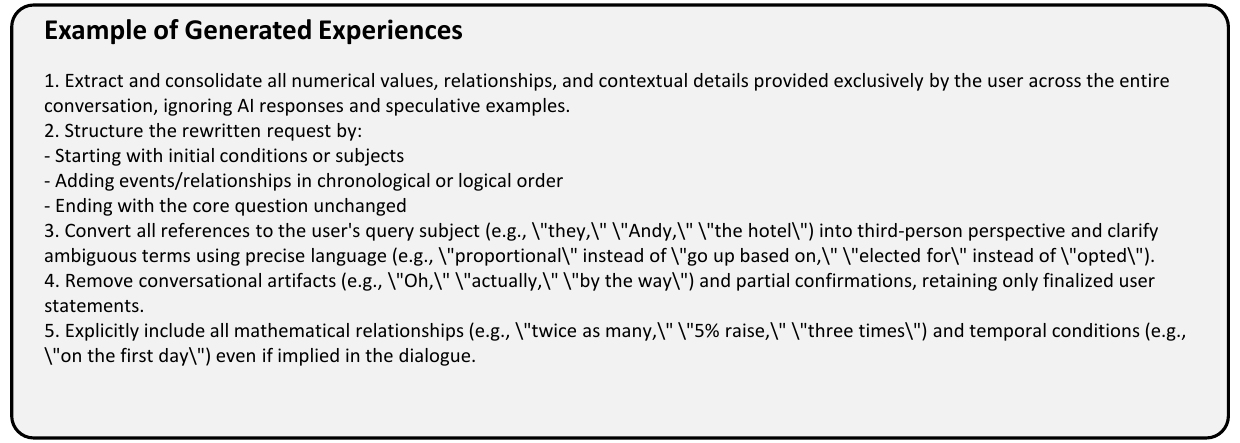}
  \label{p3}
\end{figure}

\end{document}